\pdfoutput=1
\documentclass[11pt]{article}
\PassOptionsToPackage{table}{xcolor} % added
\usepackage{EMNLP2022}
\usepackage{arydshln}
\usepackage{times}
\usepackage{latexsym}
\usepackage{booktabs}
\usepackage{float}
\usepackage{graphicx}
\usepackage{multirow}
\usepackage{ragged2e,array,booktabs}
\usepackage{graphicx}
\usepackage[export]{adjustbox}
\usepackage[T1]{fontenc}

\newcommand{\stitle}[1]{\vspace{0.3ex} \noindent{\bf #1}}
\usepackage[utf8]{inputenc}

\usepackage{microtype}

\usepackage{inconsolata}

\title{Self-supervised Graph Masking Pre-training for Graph-to-Text Generation}
\author{Jiuzhou Han \and Ehsan Shareghi \\ 
Department of Data Science \& AI, Monash University \\ 
\texttt{\{jiuzhou.han, ehsan.shareghi\}@monash.edu}}
\begin{document}
\maketitle
\begin{abstract}
Large-scale pre-trained language models~(PLMs) have advanced Graph-to-Text~(G2T) generation by processing the linearised version of a graph. However, the linearisation is known to ignore the structural information. Additionally, PLMs are typically pre-trained on free text which introduces domain mismatch between pre-training and downstream G2T generation tasks. To address these shortcomings, we propose graph masking pre-training strategies that neither require supervision signals nor adjust the architecture of the underlying pre-trained encoder-decoder model. When used with a pre-trained T5, our approach achieves new state-of-the-art results on WebNLG+2020 and EventNarrative G2T generation datasets. Our method also shows to be very effective in the low-resource setting.\footnote{Our code is available at \url{https://github.com/Jiuzhouh/Graph-Masking-Pre-training}.}
\end{abstract}

\section{Introduction}
% \begin{itemize}
%     \item Significance of Graph-to-Text generation tasks and the fact that Transformer models are now dominating this area (1 paragraph). Also highlighting why we want to use sequence encoders instead of graph encoders.
%     \item Reviewing existing graph-to-text generation approaches, highlight their need to use labelled training data, and the fact that unsupervised approaches still underperform the supervised counterparts (1 paragraph)
%     \item We propose various self-supervised pre-training Graph Masking strategies that do not require labeled data and leverage graph structure information and masked path prediction during pre-training. We illustrate how existing pre-trained T5 model could be further augmented with this training strategy towards graph-to-text generation task (1 paragraph)
%     \item Our experimental findings on graph-to-text generation across 4 datasets .... (1 paragraph)
% \end{itemize}

Graph-to-Text (G2T) generation \citep{DBLP:journals/jair/GattK18} is the task of generating natural language from graph-structured data. While there are several tasks that could leverage a G2T component~\cite{,DBLP:conf/ijcai/ZhouYHZXZ18,DBLP:conf/emnlp/JiKHWZH20,DBLP:conf/iclr/ChenCSWC21} the direct generation of text description from knowledge graphs (KGs) have attracted a lot of attention due to its potential in providing a more accessible presentation of knowledge to non-experts~\cite{schmitt-etal-2020-unsupervised}.
%\ehsan{Need to motivate this first as a G2T task itself. Then can mention the applications...} 

%It has achieved great success in many downstream NLP applications such as dialogue systems \citep{DBLP:conf/naacl/WenGMRSVY16,DBLP:conf/ijcai/ZhouYHZXZ18}, story generation \citep{DBLP:conf/aaai/GuanWH19,DBLP:conf/emnlp/JiKHWZH20} and question answering \citep{DBLP:conf/naacl/AgarwalGSA21,DBLP:conf/iclr/ChenCSWC21}. 

%The approaches proposed to solve this generation task can be broadly divided into two categories based on how to handle the graph. The first kind of method is to handle the graph directly, which uses a graph-to-sequence model \citep{DBLP:conf/inlg/MarcheggianiP18} \citep{DBLP:conf/acl/CohnHB18} \citep{DBLP:conf/inlg/HanBC21}. By using a graph encoder such as graph convolutional network \citep{DBLP:conf/iclr/KipfW17}, graph attention network \citep{DBLP:conf/iclr/VelickovicCCRLB18}, this approach can exploit the graph structure and utilise the structural information to help the text decoder to generate target sentences. Another kind of method is to use a sequence-to-sequence model \citep{DBLP:conf/acl/KonstasIYCZ17,DBLP:conf/acl/WangZQT18,DBLP:conf/emnlp/FerreiraLMK19}. The key step in this approach is to linearise the input graph to a sequence, so this approach usually ignores the graph structure.
In parallel, Transformer-based \citep{DBLP:conf/nips/VaswaniSPUJGKP17} pre-trained language models (PLMs) such as BART \citep{lewis2019bart}, and T5 \citep{raffel2019exploring} have facilitated state-of-the-art (SotA) results on several tasks, including earlier SotA results for G2T~\citep{DBLP:journals/corr/abs-2007-08426,DBLP:conf/inlg/KaleR20a,mager-etal-2020-gpt}. It has been argued that their success, in part, is due to factual memorisation that guides the generation~\cite{DBLP:journals/corr/abs-2007-08426}. Although PLMs benefit the G2T generation, the linearisation step required to use these models ignores the structural information of the graph \citep{DBLP:conf/acl/WangYLJR21}, while explicitly modelling structured data could also lead to catastrophic forgetting of distributional knowledge \citep{DBLP:conf/emnlp/RibeiroZG21}. 

To address this, \citet{DBLP:conf/acl/WangYLJR21} proposed adding extra positional embedding layers to capture the inter-dependency structures of input graphs. \citet{DBLP:conf/emnlp/RibeiroZG21} proposed using a structure-aware adapter in PLMs to  supplement the input with its graph structure. For table data, \citet{xing-wan-2021-structure} considered the structure of the table input by predicting the surrounding cells for a cell in a table. However, these methods either change the design of the PLMs (limiting their use for other task settings) or require labelled training data to capture the graph structure information.

In this work, we propose self-supervised graph masking pre-training strategies to enhance the structure awareness of PLMs. To achieve this, we formulate several graph masking strategies to inject local and global awareness of the input structure into the PLM. Our method has two key advantages: (i) it does not require to introduce extra layers or change of architecture in the underlying PLM, and (ii) it pre-trains the PLMs in a self-supervised setting on graphs, without requiring labelled training data. Starting from an existing PLM, we further pre-train it with our approach, then the fine-tuning on downstream tasks is done as per usual. 
\begin{table*}[t]
\small
    \centering
    \scalebox{0.8}{
    \begin{tabular}{>{\RaggedRight}p{2.5cm}>{\RaggedRight}p{10.5cm}>{\RaggedRight}p{5.5cm}}
    \toprule
    \textbf{Pre-training Task} & \textbf{Input (Triples format: $[\texttt{S} | \texttt{head}_1, \texttt{P} | \texttt{relation}_1, \texttt{O} | \texttt{tail}_1, l_1]$)} & \textbf{Target Output} \\
    \midrule
     Triple Prediction & [\textcolor{red}{<X>}, \texttt{1}], [\texttt{S} | New York City, \texttt{P} | country, \texttt{O} | United States, \texttt{2}], [\texttt{S} | New York City, \texttt{P} | is Part Of, \texttt{O} | Manhattan, \texttt{2}], [\texttt{S} | Manhattan, \texttt{P} | leader Name, \texttt{O} | Cyrus Vance Jr., \texttt{3}], [\texttt{S} | Manhattan, \texttt{P} | is Part Of, \texttt{O} | New York, \texttt{3}] & \textcolor{red}{<X>} [\texttt{S} | Asser Levy Public Baths, \texttt{P} | location, \texttt{O} | New York City] <Z> \\
    %\midrule
     %Triple Prediction (with removal) & [\textcolor{red}{<X>}, \texttt{1}], [\texttt{S} | New York City, \texttt{P} | country, \texttt{O} | United States, \texttt{2}], [\texttt{S} | New York City, \texttt{P} | is Part Of, \texttt{O} | Manhattan, \texttt{2}] & \textcolor{red}{<X>} [\texttt{S} | Asser Levy Public Baths, \texttt{P} | location, \texttt{O} | New York City] <Z>  \\
    \midrule
    Relation Prediction & [\texttt{S} | Asser Levy Public Baths, \texttt{P} | location, \texttt{O} | New York City, \texttt{1}], [\texttt{S} | New York City, \textcolor{blue}{<Y>}, \texttt{O} | United States, \texttt{2}], [\texttt{S} | New York City, \texttt{P} | is Part Of, \texttt{O} | Manhattan, \texttt{2}], [\texttt{S} | Manhattan, \texttt{P} | leader Name, \texttt{O} | Cyrus Vance Jr., \texttt{3}], [\texttt{S} | Manhattan, \texttt{P} | is Part Of, \texttt{O} | New York, \texttt{3}] & \textcolor{blue}{<Y>} \texttt{P} | country <Z>  \\
    \midrule
     Triple Prediction +  Relation Prediction & [\textcolor{red}{<X>}, \texttt{1}], [\texttt{S} | New York City, \texttt{P} | country, \texttt{O} | United States, \texttt{2}], [\texttt{S} | New York City, \texttt{P} | is Part Of, \texttt{O} | Manhattan, \texttt{2}], [\texttt{S} | Manhattan, \textcolor{blue}{<Y>}, \texttt{O} | Cyrus Vance Jr., \texttt{3}], [\texttt{S} | Manhattan, \texttt{P} | is Part Of, \texttt{O} | New York, \texttt{3}] &\textcolor{red}{<X>} [\texttt{S} | Asser Levy Public Baths, \texttt{P} | location, \texttt{O} | New York City] \textcolor{blue}{<Y>} \texttt{P} | leader Name <Z> \\
    \bottomrule
\end{tabular}}
%\vspace{-0.1cm}
\caption{The input-output format for our graph masking strategies.}
%\vspace{-0.4cm}
\label{tab:grask_masking}
\end{table*}
We conduct extensive experiments on three G2T generation datasets of diverse graphs. %and illustrate how an existing pre-trained T5 model could be further augmented with our training strategy. 
Our empirical findings highlight that our self-supervised strategies significantly outperform a strong underlying T5 baseline and achieve two new SotA results on two of the datasets WebNLG+2020~\citep{zhou-lampouras-2020-webnlg} and EventNarrative~\citep{DBLP:conf/nips/ColasSWW21}. Additionally, we show our pre-training strategies are very efficient in utilising data and have a great potential for low-resource setting. 
%\ehsan{punchline about what this work enables the community to achieve or develop further}
\section{Self-Supervised Graph Masking}

Our desiderata is to infuse structural knowledge into widely used pre-trained encoder-decoder Transformer models, without modifying the model architecture or relying on supervision signal. To achieve this, we propose three self-supervised learning tasks to further pre-train a \texttt{T5-LARGE} \citep{DBLP:journals/jmlr/RaffelSRLNMZLL20} model prior to fine-tuning on G2T generation downstream tasks. In this section we first describe our graph linearisation step which prepares the data in the right format for T5 encoder while injecting some weak structural information into the input~(\S\ref{sec:linearise}), then we introduce our three graph masking pre-training tasks (\S\ref{sec:graph_masking}).

\subsection{Linearising a Graph}\label{sec:linearise} 
We linearise a graph into a set of triples in the format of [subject, predicate, object], representing [head entity, relation, tail entity] for every edge in a graph. Following \citet{DBLP:conf/acl/WangYLJR21}, we prepend $\texttt{S} |$, $\texttt{P} |$, $\texttt{O} |$ tokens to further specialise each entity or relation with its role in a triple. Additionally, to provide a weak  structural signal from the graph, we also augment every triple by a level marker $l$, indicating the distance of its object entity from the root (the node that does not have a parent in the graph). This is similar to \citep{DBLP:conf/acl/WangYLJR21}, noting the key difference in that they embed the tree level using an extra layer together with other positional embeddings, but we simply augment the linearised input without adding any extra layers. The final augmented triple has the following format: $[\texttt{S} | \texttt{head entity}, \texttt{P} | \texttt{relation}, \texttt{O} | \texttt{tail entity}, l]$. For a visual example of this, see \emph{Appendix}~\ref{apendix_visual}.

\subsection{Graph Masking Pre-training Strategies}\label{sec:graph_masking}
The three self-supervised learning tasks are formulated as follows:

\stitle{Triple Prediction (\texttt{Triple}).}
For a linearised graph, on each level we randomly mask one full triple and replace it with a mask token <X>, which is then used as the target for prediction. The masked triple can be seen as a sub-graph of the original graph. This is to encourage the model to automatically identify the most relevant parts of a full graph related to each of its sub-graph. 
%In this setting we encourage the model to utilise the entire graph for correctly predicting the masked triple. 

%\stitle{Triple Prediction (with removal).} To emphasise on the connectivity of the subgraph, we remove the triples that do not have any head or tail entity in common with the masked triple from the input side. This encourages the model to elicit knowledge from the immediate neighbours of the masked triple.
%We introduce three self-supervised graph masking pre-training strategies to further pretrain a T5-large model on different task-specific graph datasets. The training data is constructed from the original graph and we create different training data based on different pre-training tasks.

%\paragraph{External Triple Prediction}
%For a linearised graph, on each tree level we randomly mask one triple with a mask token <X>. The masked triple can be seen as a subgraph of the original graph and the goal is to use other parts of the graph to predict the masked subgraph. We also add a token <Z> as the end token in the target.

%\paragraph{External Triple Prediction (with removal)}
%In External Triple Prediction strategy, we use all other triples to predict the masked triple. But in External Triple Prediction (with removal) strategy, we further consider the connectivity of the subgraph. We remove the triples that do not have connections with the masked triple and only use the triples that have connections with the masked triple (i.e. with common subject or object) to predict the masked triple.

\stitle{Relation Prediction (\texttt{Relation}).}
In this strategy, we focus on the relations within triples. We randomly mask one relation on each level with a mask token <Y>, and the model is tasked to predict the masked relation as the target. This task requires the model to leverage very local information (i.e., between a head and a tail) to predict the masked relation. Local cohesiveness is expected to translate into better translation of triples into text fragments.

\stitle{Triple + Relation Prediction (\texttt{Triple+Relation}).}
This ultimate strategy combines both Triple and Relation Prediction tasks to leverage the benefits of both worlds. In this setting,  the Triple Prediction task follows the same protocol as stated above, but for Relation Prediction, we only consider the relation in triples that are not connected with the masked triple. We randomly mask one triple with the mask token <X>. For the triples that do not have common subject or object with the masked triple, we also randomly mask one relation with the mask token <Y>. The model jointly learns to predict both the masked sub-graphs and relations at the same time. 

In all pre-training tasks we also add a token <Z> as the end token in the target output. Table \ref{tab:grask_masking} summarises these three pre-training tasks via an example of each kind of graph masking strategy. Graph Masking Pre-training follows the standard cross-entropy loss, which is to minimise the negative log-likelihood of the masked part of the graph:
$$
\mathcal{L}_{GMP}=-\sum_{i=1}^N \log p\left(m_i \mid x_i\right)
$$
where $m_i$ is the masked part of the graph, $x_i$ is the unmasked part of the graph, $N$ is the number samples.

%By further training the pretrained language on the External Triple Prediction task and Internal Relation Prediction task, we encourage the model to learn some local information about the relation in the triple and some global information between the triples in the subgraph level, respectively.

\section{Experiments}

% \begin{itemize}
%     % \item Overview and statistics of WebNLG, DART, Event Narrative tasks and datasets
%     % \item T5 model, task finetuning, and training configurations
%     % \item evaluation metrics (automatic on full test set, and human evaluation on 50 cases?)
% \end{itemize}

% We evaluate on four standard graph-to-text generation datasets: WebNLG 2017 \citep{DBLP:conf/inlg/GardentSNP17}, WebNLG+2020, DART \citep{DBLP:conf/naacl/NanRZRSHTVVKLIP21}, AGENDA \citep{DBLP:conf/naacl/Koncel-Kedziorski19}, EventNarrative \citep{DBLP:conf/nips/ColasSWW21}. WebNLG 2017 dataset contains a set of triples extracted from DBpedia \citep{DBLP:conf/semweb/AuerBKLCI07} and some texts to describe these triples for 15 distinct DBpedia categories. WebNLG+2020 is similar to WebNLG 2017 but with new data of a new DBpedia category. DART is an open-domain structured dataset collected from different sources which cover a broad range of topics. AGENDA contains scientific abstracts extracted from proceedings of AI conferences paired with paper title and automatically generated knowledge graphs. EventNarrative is a large-scale, event-centric dataset extracted and paired from existing large-scale data repositories, including Wikidata, Wikipedia, and EventKG \citep{DBLP:conf/esws/GottschalkD18}.
In this section we outline the experimental setups~(\S\ref{sec:expset}), followed by downstream G2T generation results in full (\S\ref{sec:results}) and low-resource scenarios (\S\ref{sec:lr_results}). We also present a set of generated outputs from our models (\S\ref{sec:generation}), and finish by providing an analysis~(\S\ref{sec:analysis}) on the effect of pre-training data size, and an ablation on the role of input augmentation with level markers.
\subsection{Experimental Setups}\label{sec:expset}
\stitle{Tasks and Datasets.}
We evaluate on three G2T generation datasets: WebNLG+2020 \citep{zhou-lampouras-2020-webnlg}, DART \citep{DBLP:conf/naacl/NanRZRSHTVVKLIP21}, EventNarrative \citep{DBLP:conf/nips/ColasSWW21}. WebNLG+2020\footnote{\url{https://gitlab.com/shimorina/webnlg-dataset/-/tree/master/release_v3.0}} contains a set of triples extracted from DBpedia~\citep{DBLP:conf/semweb/AuerBKLCI07} and text description for 16 distinct DBpedia categories. DART\footnote{\url{https://github.com/Yale-LILY/dart}} is an open-domain heterogeneous structured dataset collected from different sources which cover a broad range of topics. EventNarrative\footnote{\url{https://www.kaggle.com/datasets/acolas1/eventnarration}} is a large-scale, event-centric dataset extracted and paired from existing large-scale data repositories, including Wikidata, Wikipedia, and EventKG~\citep{DBLP:conf/esws/GottschalkD18}. See \emph{Appendix}~\ref{appendix_data} for full data statistics. 
%Table \ref{table:stat}.

\stitle{Pre-training Datasets.} For each pre-training strategy, we create the pre-training datasets on the graph side of the task training data with the right format. 
%WebNLG+2020 and DART. For EventNarrative, since all graphs are 1-hop graphs which means all triples have the common root entity node, we do not have Triple Prediction (with removal) pre-training task.}

\stitle{Evaluation Metrics.} We report the automatic evaluation using \texttt{BLEU}~\citep{DBLP:conf/acl/PapineniRWZ02}, \texttt{METEOR}~\citep{DBLP:conf/acl/BanerjeeL05}, \texttt{TER}~ \citep{DBLP:conf/amta/SnoverDSMM06} which are used in the official WebNLG challenge~ \citep{DBLP:conf/inlg/GardentSNP17} and \texttt{BERTScore}~ \citep{DBLP:conf/iclr/ZhangKWWA20} which considers the semantic meanings of words or phrases.

%\stitle{Implementation Details.} Our implementation is based on the Huggingface Library \citep{DBLP:journals/corr/abs-1910-03771}. Optimisation was done using Adam~\citep{DBLP:journals/corr/KingmaB14} with a learning rate of 3e-5 and a batch size of 3 both in the pre-training and fine-tuning stages. We use a single NVIDIA V100 GPU for all experiments and fine-tuning experiments took from 20 to 100 hours depending on the size of dataset.
\begin{table}[t]
\setlength{\tabcolsep}{3pt}
\centering
\scalebox{0.75}{
\begin{tabular}{lcccccc}
\toprule
\bf \parbox[t]{2mm}{\multirow{1}{*}{\rotatebox[origin=c]{90}{Data}}}&\multicolumn{1}{p{1.5cm}}{\centering  Metric}& \multicolumn{1}{p{1cm}}{\centering \texttt{T5} \texttt{LARGE}} & \multicolumn{1}{p{1cm}}{\centering  \texttt{SotA}}  &  \multicolumn{1}{p{1.5cm}}{\centering  \texttt{Triple}} &   \multicolumn{1}{p{1.5cm}}{\centering  \texttt{Relation}} & \multicolumn{1}{p{2cm}}{\centering \texttt{Triple+}  \texttt{Relation}}\\
% \bf \parbox[t]{2mm}{\multirow{1}{*}{\rotatebox[origin=c]{90}{Data}}}&\multicolumn{1}{p{1.5cm}}{\centering \\ Metric}& \multicolumn{1}{p{1cm}}{\centering \texttt{T5}\\ \texttt{LARGE}} & \multicolumn{1}{p{1cm}}{\centering \\ \texttt{SotA}}  &  \multicolumn{1}{p{1.5cm}}{\centering \\ \texttt{Triple}} &   \multicolumn{1}{p{1.5cm}}{\centering \\ \texttt{Relation}} & \multicolumn{1}{p{2cm}}{\centering \texttt{Triple+} \\ \texttt{Relation}}\\
 \midrule
 \bf \parbox[t]{2mm}{\multirow{4}{*}{\rotatebox[origin=c]{90}{WebNLG}}} &\texttt{BLEU}& 53.60 & 55.41 & 
 \bf{57.64} & 56.93 & 57.49 \\
 &\texttt{METEOR} & 39.52 & 41.90 & \bf{42.24} & 41.94 & 42.19 \\
 &\texttt{TER} &41.48 & 39.20 & \bf{38.86} & 39.42 & 39.08 \\
 &\texttt{BERTScore} &95.02 & - & \bf{95.36} & 95.23 & 95.28 \\
  \midrule
\bf \parbox[t]{2mm}{\multirow{4}{*}{\rotatebox[origin=c]{90}{EventNar}}} &\texttt{BLEU}& 34.31 & 35.08 & 38.27 & \bf{38.36} & 38.08 \\
 &\texttt{METEOR}& 26.84 & 27.50 & \bf{31.01} & 30.80 & 30.99 \\
 &\texttt{TER} &58.26 & - & \bf{55.19} & 56.11 & 55.32 \\
 &\texttt{BERTScore} &93.02 & 93.38 & \bf{95.24} & 95.07 & 95.21 \\
 \midrule
 \bf \parbox[t]{2mm}{\multirow{4}{*}{\rotatebox[origin=c]{90}{DART}}} &\texttt{BLEU}& 50.66 & \bf{51.95} & 50.85 &50.71 & 50.83 \\
 &\texttt{METEOR}& 40 & \bf{41.07} & 40.31 & 40.23 & 40.37 \\
 &\texttt{TER} &43 & \bf{42.75} & 43.23 & 43.68 & 43.51 \\
 &\texttt{BERTScore} &95 & 95 & 95.11 & 95.04 & 95.16 \\
 \bottomrule
\end{tabular}
}
\caption{G2T generation results on 3 datasets.}
\label{table:results}
%\vspace{-0.2cm}
\end{table}
\stitle{Baseline, SotA, Our Models.} We use the \texttt{T5-LARGE} model as our baseline for fine-tuning. T5-large results are based on the published results~\cite{DBLP:journals/corr/abs-2007-08426}. All our models further pre-train the vanilla \texttt{T5-LARGE} model and are further fine-tuned for G2T generation tasks as usual. We denote our configurations as \texttt{Triple}, \texttt{Relation}, \texttt{Triple+Relation}. SotA results for WebNLG and DART are from \citet{DBLP:journals/corr/abs-2110-08329}, and for EventNarrative are based on \citet{DBLP:journals/corr/abs-2204-06674}. Our implementation is based on the Huggingface Library \citep{DBLP:journals/corr/abs-1910-03771}. Optimisation was done using Adam~\citep{DBLP:journals/corr/KingmaB14} with a learning rate of 3e-5 and a batch size of 3 both in the pre-training and fine-tuning stages. We used a V100 16GB GPU for all experiments.

\subsection{Graph-to-text Generation}\label{sec:results}
\stitle{Task Formulation.} G2T generation follows the standard language modelling objective. Given an input graph $\mathcal{G}$, the model aims to generate ground-truth text $y=(y_1, ... , y _N)$. The objective is to maximise the likelihood of the ground-truth text, which is equivalent to minimise the negative log-likelihood as:
$$
\mathcal{L}_{G2T}=-\sum_{i=1}^N \log p\left(y_i \mid y_1, \ldots, y_{i-1} ; \mathcal{G}\right)
$$

\stitle{Results.} Table \ref{table:results} reports the results of fine-tuning the baseline, SotA and our models on three G2T generation tasks. For WebNLG, all of our strategies outperform both the baseline and SotA results. The performance difference among our three variants is statistically insignificant. Similarly, on EventNarrative all our models outperform SotA and baseline. For DART, the improvement over the baseline is not as significant as for the other two datasets, while our method matches SotA on BERTScore but falls behind on the other metrics. We speculate this to be reflective of the heterogeneous nature of DART, which has a large proportion of data with very limited relations (e.g., roughly 52\% of DART contains only 7 types of relations). In this setting, the pre-training tasks cannot capture much useful structure information on this sparse data.

\begin{table}[t]
\scalebox{0.58}{
\centering
\begin{tabular}{llcccc}
\hline
Tr.Size & Model Setting & \texttt{BLEU} & \texttt{METEOR} & \texttt{TER} & \texttt{BERTScore}  \\ \hline
\multirow{3}{*}{5\%} 
 & w/o pre-training & 48.52 & 37.44 & 43.97 & 94.66 \\
 & same 5\% for pre-training  & \bf{52.79} & \bf{40.41} & \bf{42.02} & \bf{94.94} \\
 & remaining 95\% for pre-training & 50.69 & 39.06 & 42.97 & 94.72 \\ \hline
\multirow{3}{*}{10\%} 
 & w/o pre-training & 48.64 & 37.24 & 43.33 & 94.65 \\
 & same 10\% data for pre-training & \bf{53.56} & \bf{40.45} & \bf{41.19} & \bf{95.03} \\
 & remaining 90\% data for pre-training & 52.57 & 39.75 & 42.17 & 94.75 \\ \hline
\multirow{3}{*}{25\%} 
 & w/o pre-training & 50.35 & 37.87 & 43.82 & 94.66 \\
 & same 25\% data for pre-training & \bf{56.04} & \bf{41.57} & \bf{39.38} & \bf{95.24} \\
 & remaining 75\% data for pre-training & 55.93 & 41.46 & 39.78 & 95.20 \\ \hline
\end{tabular}}
\caption{Results of each model in the low-resource setting on WebNLG+2020 dataset. Tr.Size denotes the amount of data used for downstream task fine-tuning.}
\label{table:low-data}
\vspace{-0.2cm}
\end{table}

\subsection{Low-resource Setting}\label{sec:lr_results}
We investigated the performance of our methods in low-resource scenario. For this we used \texttt{Triple} as the pre-training strategy and k\% (k=5, 10, 25) of WebNLG+2020 training data for downstream task fine-tuning. We tried two configurations to see if pre-training (still without using the labels) with the same training data would be better than pre-training on the non-overlapping training data: (1) used the same k\% between pre-training and fine-tuning, (2) used 100-k\% for pre-training and k\% for fine-tuning. We compared the results of these two settings with the \texttt{T5 LARGE} which was only task fine-tunined (without additional pre-training). The results are shown in Table \ref{table:low-data}. 

The models using pre-training significantly outperform the models without pre-training. For instance in 5\% training scenario, the pre-trained model with \texttt{Triple} which used the same amount of data for both pre-training and fine-tuning outperforms \texttt{T5 LARGE} by a margin of 4 BLEU scores. This indicates that our graph masking pre-training strategies can effectively improve the performance of the underlying PLM in the low-resource scenario. With the increment of training data, the improvement effect of pre-training method is greater. Moreover, pre-training with the same training data leads to better results compared with using non-overlapping data. We speculate this happens since the model in this configuration is exposed to learn specific structural knowledge that will be used in the seen training data for fine-tuning downstream tasks. This also suggests a potential for our approach in multi-task learning, which we leave to future work. As the increase of training data, the gap of the performance of pre-training using different parts of data also decreases.

\begin{table}[t]
\centering
\scalebox{0.85}{
\begin{tabular}{lccccc}
\hline
Data & Time & \texttt{BLEU} & \texttt{METEOR} & \texttt{TER} & \texttt{BERTScore}  \\
\hline
100\%  & 10h    & 57.64    & 42.24  & 38.86   & 95.36  \\
75\%  & 7.5h   & 56.92    & 41.98  & 39.07   & 95.29     \\
50\%  & 5h   & 56.78     & 41.96   & 39.90 & 95.18    \\
25\%  & 2.5h   & 56.73     & 41.85   & 40.13 & 95.21    \\
5\%  & 0.5h   & 56.40    & 41.28  & 40.24   & 95.12    \\
\hline
\end{tabular}}
\caption{Results of using different amounts of pre-training data in \texttt{Triple} strategy on WebNLG+2020. Time denotes the pre-training duration.}
\label{table:datasize change}
\end{table}
\subsection{Generated Samples}\label{sec:generation}
We demonstrate two qualitative examples of generated texts on WebNLG+2020 and EventNarrative test sets in Table \ref{tab:samples}. 

For the WebNLG example, while \texttt{T5 LARGE} generates fluent texts but misses to cover the ``recorded in'' relation. Previous SotA model generates all information from the graph, but it breaks the order of arguments for ``preceded By''. While our model can not only produce the sentences with correct information. 

For the EventNarrative example, the ``Russian'' information in the reference does not exist in the graph, which should be inferred by the PLM. For \texttt{T5 LARGE} and previous SotA, neither can generate such information, while our model can generate this additional information without missing any information from the graph. See more generated samples in \emph{Appendix}~\ref{apendix_generation}.

\begin{table*}[t]
    \centering
    \scalebox{0.94}{
    \begin{tabular}{>{\RaggedRight}p{8cm}>{\RaggedRight}p{8cm}}
    \toprule
    \textbf{WebNLG+2020}  & \textbf{EventNarrative} \\
    \midrule
    \includegraphics[scale=0.28]{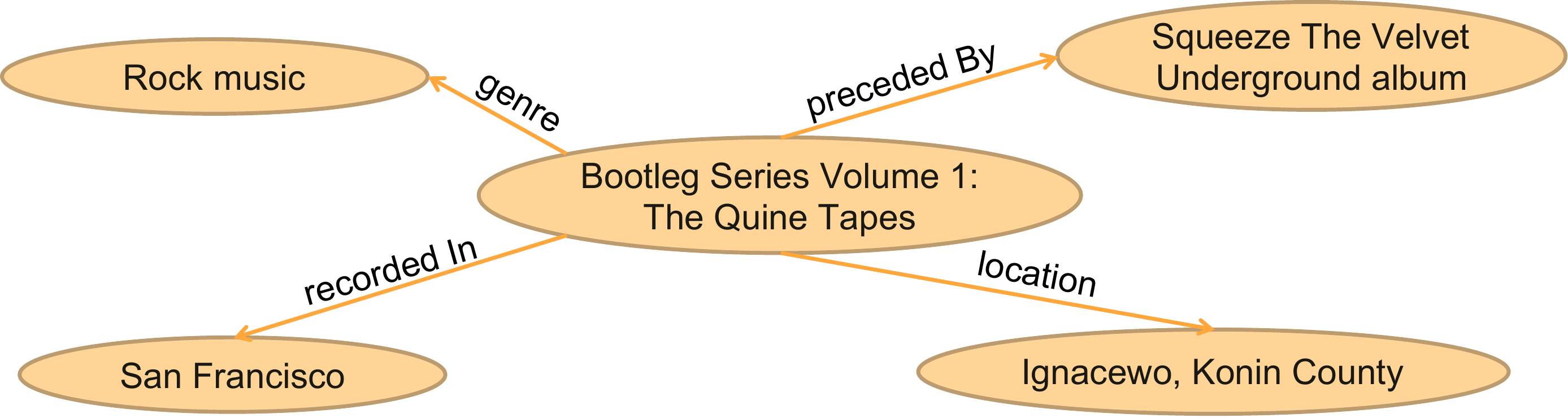}  & \includegraphics[scale=0.3]{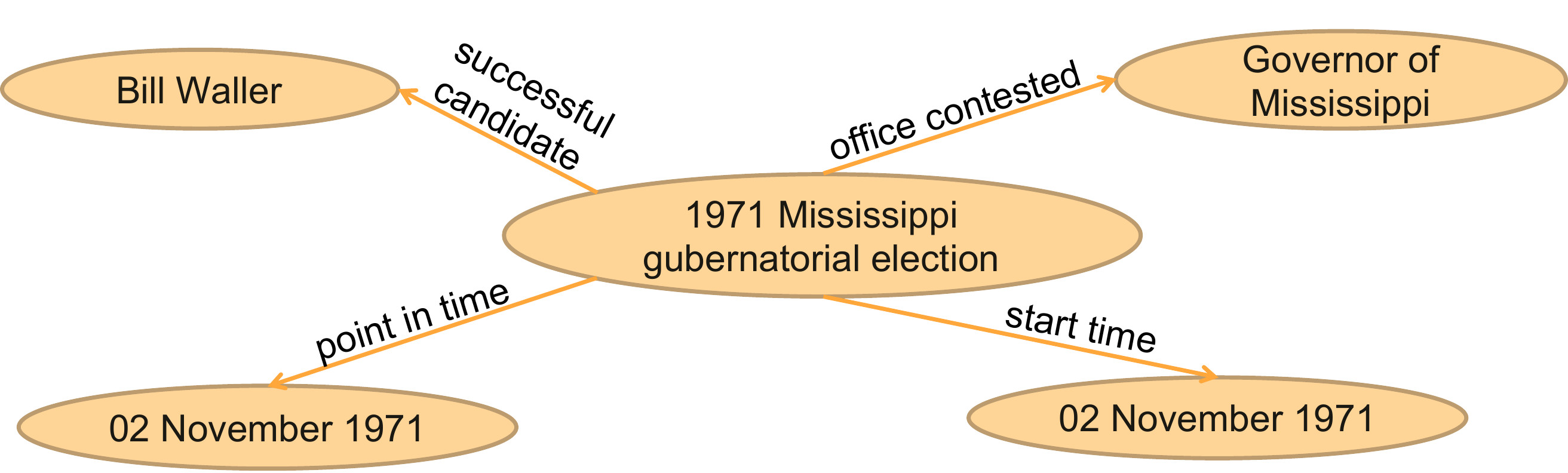} \\
    \midrule
    \textbf{Reference:} The Velvet Underground Squeeze album was succeeded by the rock album Bootleg Series Volume 1: The Quine Tapes, recorded under record label Polydor Records in San Francisco.  & \textbf{Reference:} The First Battle of Ignacewo was one of many clashes of the January Uprising. it took place on may 8, 1863, near the village of Ignacewo, Konin County, which at that time belonged to Russian empire’s Congress Poland.  \\
    \midrule
   \textbf{T5-Large:} The genre of Bootleg Series Volume 1: The Quine Tapes is rock music and was preceded by the album Squeeze The Velvet Underground. The album was released by Polydor Records.   & \textbf{T5-Large:} The First Battle of Ignacewo was fought in Ignacewo, Konin County, Congress Poland, during the January Uprising.  \\
    \midrule  
    \textbf{Previous SotA:} Squeeze The Velvet Underground was preceded by Bootleg Series Volume 1: The Quine Tapes, which was recorded in San Francisco and released by Polydor Records. The genre of the album is rock music.   &  \textbf{Previous SotA:} The First Battle of Ignacewo was one of the ﬁrst battles of the January Uprising. It took place on January 6, 1863, near the village of Konin, in Congress Poland.  \\
    \midrule
      \textbf{Graph Masking Pre-training+T5-Large:} Bootleg Series Volume 1: The Quine Tapes, whose genre is rock music, were recorded in San Francisco and are signed to Polydor Records. They were preceded by the album Squeeze The Velvet Underground.  & \textbf{Graph Masking Pre-training+T5-Large:} The First Battle of Ignacewo was one of battles of the January Uprising. It took place on January 11, 1863, near the village of Ignacewo, Konin County, Russian-controlled Congress Poland.\\
\bottomrule
\end{tabular}}
%\vspace{-0.1cm}
\caption{Examples of output texts on WebNLG+2020 and EventNarrative test sets.}
%\vspace{-0.4cm}
\label{tab:samples}
\end{table*}

\subsection{Analysis}\label{sec:analysis}
\stitle{Effect of Pre-training Data Size.} To explore how the size of the used pre-training data affects the performance of our strategies in downstream tasks, we experimented on WebNLG+2020 dataset using our \texttt{Triple} strategy. We used 5\%, 10\%, 25\%, 50\%, and 100\% of the graph side of training data for pre-training, and the whole training data to fine-tune the models. We recorded the performance, and training duration in Table \ref{table:datasize change}. As the amount of pre-training data decreased, the performance of the model also decreased slightly. However, even with using 5\% of pre-training data and less than 30 minutes spent on pre-training, our method outperforms both the SotA and \texttt{T5 LARGE} models (Table~\ref{table:results}) by a significant margin.

%even with only using 5\% of pre-training data, our method still can achieve over 95\% performance comparing with using all the pre-training data and the pre-training time is also shortened $20\times$. This shows that our graph masking pre-training strategy is quite efficient.

\begin{table}[t]
\centering
\scalebox{0.77}{
\begin{tabular}{lcccc}
\hline
Pre-training Tasks   & \texttt{BLEU} & \texttt{METEOR} & \texttt{TER} & \texttt{BERTScore}  \\
\hline
\texttt{Triple}      & 57.64    & 42.24  & 38.86   & 95.36  \\
-w/o level marker     & 56.48    & 41.77  & 39.94   & 95.17     \\
\hline
\texttt{Triple+Relation}& 57.49     & 42.19   & 39.08 & 95.28    \\
-w/o level marker      & 56.28    & 41.70  & 39.72   & 95.24    \\
\hline
No pre-training      & 54.86    & 40.62  & 40.58   & 95.09    \\
-w/o level marker      & 53.60    & 39.52  & 41.48   & 95.02    \\
\hline
\end{tabular}}
\caption{Ablation results on WebNLG+2020 dataset.}
\label{table:w/o tree level}
\end{table}
\stitle{Ablation.}
To show the contribution of input augmentation with level markers (\S\ref{sec:linearise}), we experimented with \texttt{Triple} and \texttt{Triple+Relation} strategies on WebNLG+2020. We also report the results of using input augmentation with level marker during fine-tuning \texttt{T5 LARGE}. The results are shown in Table \ref{table:w/o tree level}. We observe that the input augmentation with level markers brings improvement across all settings, even when it is only used during fine-tuning (last two rows of Table~\ref{table:w/o tree level}). We speculate this to be an indication that some useful positional information is augmented to the the linearised input through adding level markers.

\section{Conclusion and Future Work}

We proposed various self-supervised pre-training strategies to improve the structural awareness of PLMs without refining the architecture or relying on labelled data. Our graph masking  strategies outperformed the strong PLM baseline and achieve new state-of-the-art results on WebNLG+2020 and EventNarrative datasets. We demonstrated that our approach is very efficient in utilising even a small pre-training or fine-tuning datasets. For future work, we will explore different graph masking strategies to adapt for different domains of graph.

\section{Limitations}
Since our method leverages the knowledge learned by pretrained language models, it is much more effective for use in scenarios where, unlike AMR graphs, the relations inside the graph correspond to meaningful words or morphemes. Additionally, we observed our method not to work well for the cases, like in E2E \citep{DBLP:conf/inlg/DusekHR19}, that the number of relations or entities are quite sparse.

%EMNLP 2022 requires all submissions to have a section titled ``Limitations'', for discussing the limitations of the paper as a complement to the discussion of strengths in the main text. This section should occur after the conclusion, but before the references. It will not count towards the page limit.  

%The discussion of limitations is mandatory. Papers without a limitation section will be desk-rejected without review. ARR-reviewed papers that did not include ``Limitations'' section in their prior submission, should submit a PDF with such a section together with their EMNLP 2022 submission.

%While we are open to different types of limitations, just mentioning that a set of results have been shown for English only probably does not reflect what we expect. 
%Mentioning that the method works mostly for languages with limited morphology, like English, is a much better alternative.
%In addition, limitations such as low scalability to long text, the requirement of large GPU resources, or other things that inspire crucial further investigation are welcome.

\section{Ethics Statement}
Our model utilises existing pretrained language models and as such it could inherit the same ethical concerns involving these models - which are being discussed widely in the community. Our pretraining method itself does not exacerbate this issue.
%Scientific work published at EMNLP 2022 must comply with the \href{https://www.aclweb.org/portal/content/acl-code-ethics}{ACL Ethics Policy}. We encourage all authors to include an explicit ethics statement on the broader impact of the work, or other ethical considerations after the conclusion but before the references. The ethics statement will not count toward the page limit (8 pages for long, 4 pages for short papers).

\bibliography{emnlp2022}
\bibliographystyle{acl_natbib}
\appendix
\section*{Appendix}

\section{Level Marker  Augmentation}\label{apendix_visual}
\begin{figure*}[t]
    \centering
    \includegraphics[scale=0.5]{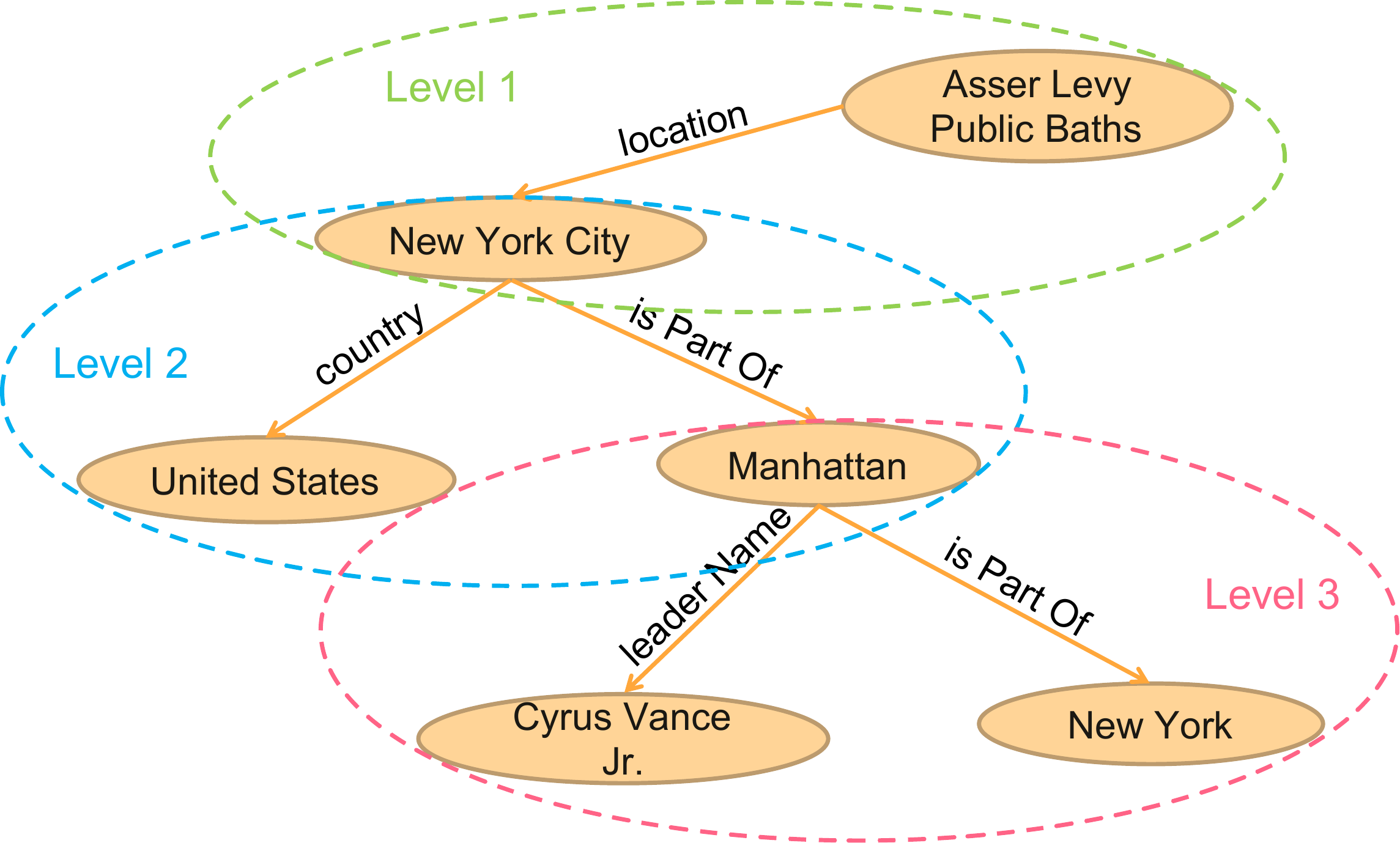}
    % \caption{An example of graph with tree Level. This graph can be represented in a linearised way: [Asser Levy Public Baths, location, New York City], [New York City, country, United States], [New York City, is Part Of, Manhattan], [Manhattan, leader Name, Cyrus Vance Jr.], [Manhattan, is Part Of, New York].}
    \caption{An example of graph with level markers. The structure-aware input of this graph is: [Asser Levy Public Baths, location, New York City, 1], [\texttt{S} | New York City, \texttt{P} | country, \texttt{O} | United States, 2], [\texttt{S} | New York City, \texttt{P} | is Part Of, \texttt{O} | Manhattan, 2], [\texttt{S} | Manhattan, \texttt{P} | leader Name, \texttt{O} | Cyrus Vance Jr., 3], [\texttt{S} | Manhattan, \texttt{P} | is Part Of, \texttt{O} | New York, 3].}
    \label{fig:tree_level}
\end{figure*}

A graph and linearised version of a level-augmented input is provided in Figure~\ref{fig:tree_level}.

\section{Data Statistics}\label{appendix_data}
The data statistics for tasks used in the paper are summarised in Table~\ref{table:stat}. 
\begin{table*}[t]
\centering
\scalebox{1.0}{
\begin{tabular}{llccccc}
\hline
Dataset & Domain & Examples & Train/Dev/Test \\ \hline
WebNLG+2020 &16 DBpedia Categories & 38,872 & 35,426/1,667/1,779 \\ \hline
EventNarrative & Events & 224,428 & 179,544/22,442/22,442 \\ \hline
\multirow{3}{*}{DART} 
 & Wikipedia& 11,998& \\
 &  15 DBpedia Categories & 27,731&62,659/6,980/12,552\\
 &  Restaurant and Hotel Descriptions  & 42,462&\\ \hline
\end{tabular}}
\caption{Statistics of WebNLG+2020, EventNarrative and DART.}
\label{table:stat}
\end{table*}

\section{Generated Samples}\label{apendix_generation}
Table \ref{more_cases} illustrates two qualitative examples of generated texts on WebNLG+2020 and EventNarrative test sets.

For the WebNLG example, \texttt{T5 LARGE} misses to cover the ``manufacturer'' and ``body Style'' information. Although previous SotA and our model both can generate correct sentences, the output of our model shows a more complex syntactic structure. For the EventNarrative example, the sentences generated from \texttt{T5 LARGE} have a big difference with the reference sentences and do not cover all information from the graph. Previous SotA model misses to cover the ``office contested'' information, while the output from our model covers all information.

\begin{table*}[t]
\centering 
\begin{tabular}{p{0.99\textwidth}}
\hline 
\Centering\textbf{WebNLG+2020} \\
\hline 
\begin{minipage}{1.0\textwidth}
    \centering 
    \includegraphics[width=0.9\textwidth]{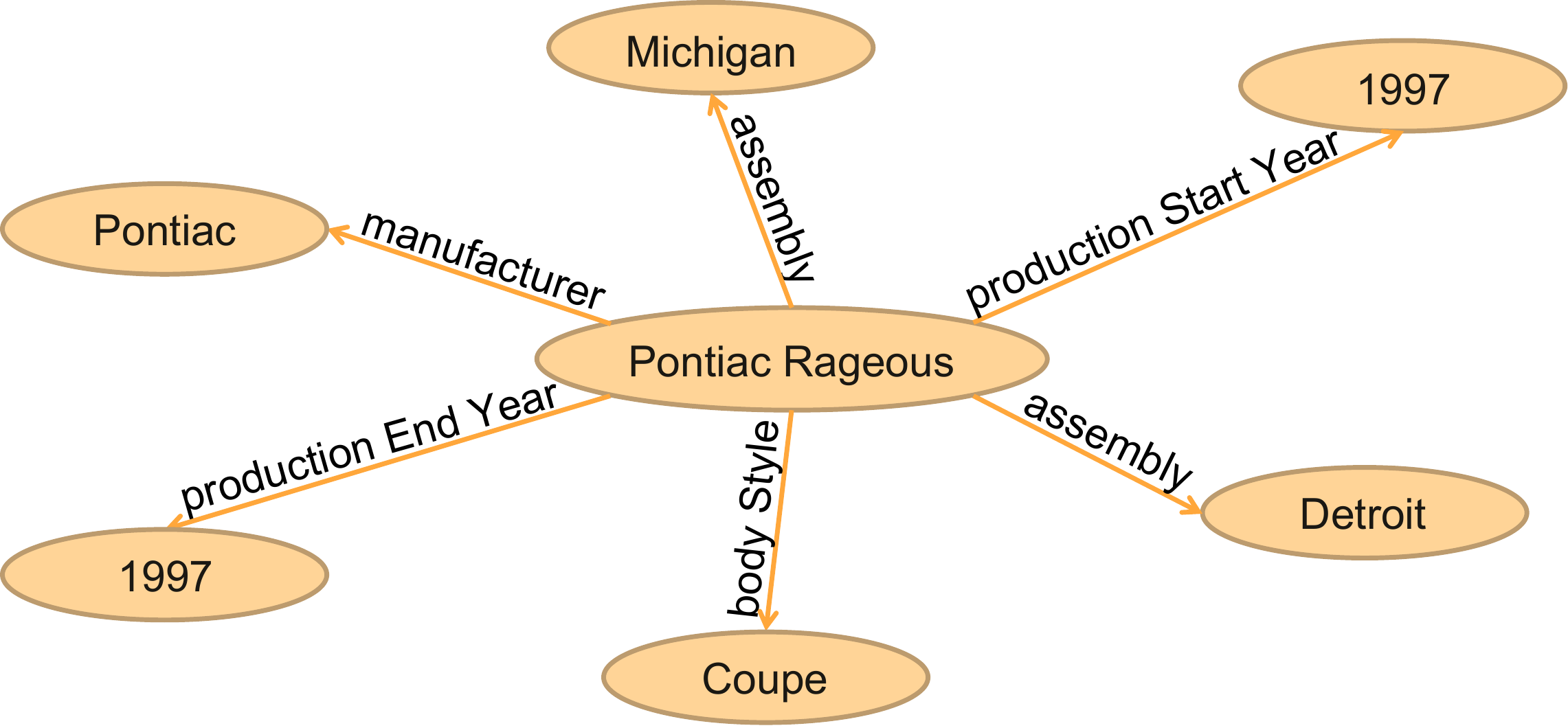}
\end{minipage} \\
\hline 
\textbf{Reference:} The Pontiac Rageous was a car with a coupe body style manufactured by Pontiac. Assembled in both Michigan and Detroit, it went into production in 1997, ending in the same year. \\
\hline 
\textbf{T5-Large:} The Pontiac Rageous is assembled in Detroit, Michigan. Its production began in 1997 and ended in 1997. The Pontiac Rageous is a 4 door, 5 passenger vehicle. \\
\hline 
\textbf{Previous SotA:} The Pontiac Rageous is manufactured by Pontiac in Detroit, Michigan. Its production began in 1997 and ended in 1997. The Pontiac Rageous has a coupe body style. \\
\hline 
\textbf{Graph Masking Pre-training+T5-Large:} Pontiac is the manufacturer of the Pontiac Rageous which has a coupe body style. The Pontiac Rageous is assembled in Detroit, Michigan and began production in 1997. \\
\hline
\hline 
\Centering\textbf{EventNarrative} \\
\hline 
\begin{minipage}{1.0\textwidth}
    \centering 
    \includegraphics[width=0.9\textwidth]{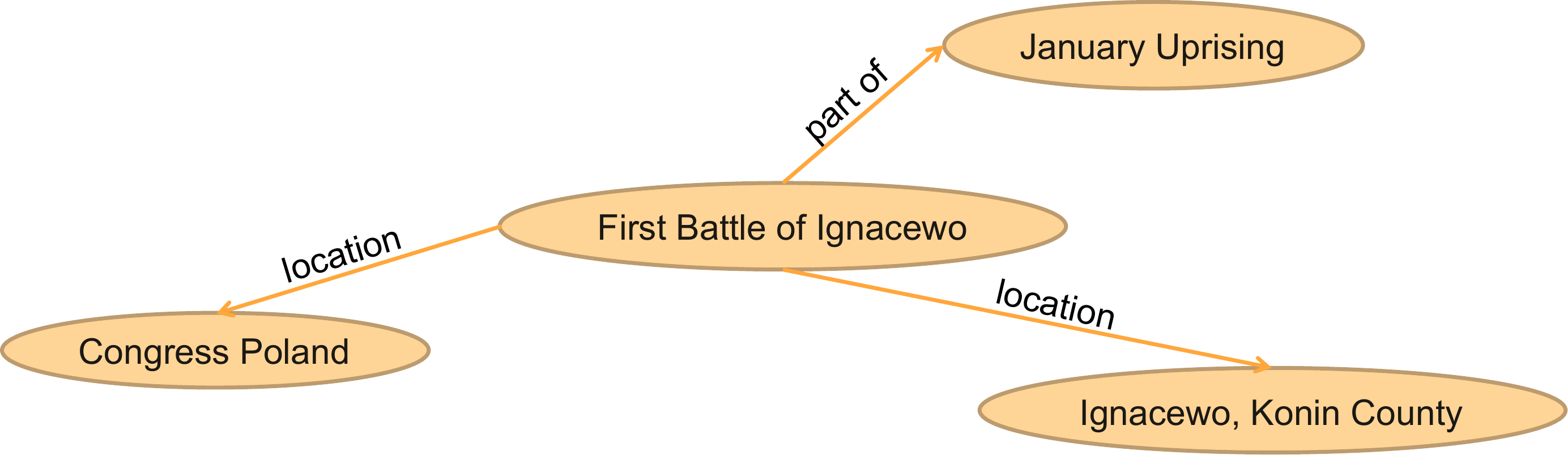}
\end{minipage} \\
\hline 
\textbf{Reference:} The First Battle of Ignacewo was one of many clashes of the January Uprising. it took place on may 8, 1863, near the village of Ignacewo, Konin County, which at that time belonged to Russian empire’s Congress Poland. \\
\hline 
\textbf{T5-Large:} The First Battle of Ignacewo was fought in Ignacewo, Konin County, Congress Poland, during the January Uprising. \\
\hline 
\textbf{Previous SotA:} The First Battle of Ignacewo was one of the ﬁrst battles of the January Uprising. It took place on January 6, 1863, near the village of Konin, in Congress Poland. \\
\hline 
\textbf{Graph Masking Pre-training+T5-Large:} The First Battle of Ignacewo was one of battles of the January Uprising. It took place on January 11, 1863, near the village of Ignacewo, Konin County, Russian-controlled Congress Poland. \\
\hline 
\end{tabular}
\caption{Examples of output texts on WebNLG+2020 and EventNarrative test sets.}
\label{more_cases}
\end{table*}

\end{document}